\documentclass[sigconf]{acmart}
\usepackage{algorithm}
\usepackage{algorithmic}
\usepackage{booktabs}
\usepackage{enumitem}
\usepackage{multirow}
\usepackage{multicol}
\usepackage{array}
\usepackage{pifont}
\newcommand{\cmark}{\ding{51}} 
\newcommand{\xmark}{\ding{55}} 

\usepackage{amsmath}
\usepackage{color}
\usepackage{xcolor}

\usepackage{listings}
\AtBeginDocument{%
  }

\setcopyright{acmlicensed}
\copyrightyear{2018}
\acmYear{2018}
\acmDOI{XXXXXXX.XXXXXXX}

\acmConference[Conference acronym 'XX]{Make sure to enter the correct
  conference title from your rights confirmation emai}{June 03--05,
  2018}{Woodstock, NY}
\acmISBN{978-1-4503-XXXX-X/18/06}

\begin{document}

\title{ECKGBench: Benchmarking Large Language Models in E-commerce Leveraging Knowledge Graph}



\author{Langming Liu\textsuperscript{1}, Haibin Chen\textsuperscript{1}, Yuhao Wang\textsuperscript{2}, Yujin Yuan\textsuperscript{1}, Shilei Liu\textsuperscript{1}, Wenbo Su\textsuperscript{1}, Xiangyu Zhao\textsuperscript{2}, Bo Zheng\textsuperscript{1}}
\affiliation{%
  \institution{\textsuperscript{1}Taobao \& Tmall Group of Alibaba, \textsuperscript{2}City University of Hong Kong}
  \country{}
}
  




  



\renewcommand{\shortauthors}{Langming Liu et al.}

\begin{abstract}
Large language models (LLMs) have demonstrated their capabilities across various NLP tasks. Their potential in e-commerce is also substantial, evidenced by practical implementations such as platform search, personalized recommendations, and customer service.
One primary concern associated with LLMs is their factuality (e.g., hallucination), which is urgent in e-commerce due to its significant impact on user experience and revenue. 
Despite some methods proposed to evaluate LLMs' factuality, issues such as lack of reliability, high consumption, and lack of domain expertise leave a gap between effective assessment in e-commerce.
To bridge the evaluation gap, we propose ECKGBench, a dataset specifically designed to evaluate the capacities of LLMs in e-commerce knowledge. 
Specifically, we adopt a standardized workflow to automatically generate questions based on a large-scale knowledge graph, guaranteeing sufficient reliability.
We employ the simple question-answering paradigm, substantially improving the evaluation efficiency by the least input and output tokens. Furthermore, we inject abundant e-commerce expertise in each evaluation stage, including human annotation, prompt design, negative sampling, and verification.
Besides, we explore the LLMs' knowledge boundaries in e-commerce from a novel perspective. 
Through comprehensive evaluations of several advanced LLMs on ECKGBench, we provide meticulous analysis and insights into leveraging LLMs for e-commerce. 
The benchmark dataset is available online at~\url{https://github.com/ming429778/ECKGBench}.
\end{abstract}



\maketitle

\section{Introduction}

The rapid growth of large language models (LLMs) has drawn significant attention in natural language processing (NLP) tasks, including but not limited to single-turn and multi-turn dialogues, semantic understanding, and text classification~\cite{OpenAI2023GPT4,touvron2023llama,chowdhery2023palm}. Equipped with powerful worldwide knowledge and generalization capabilities, LLMs demonstrate their potential in e-commerce, including various scenarios, such as platform search~\cite{guu2020retrieval}, personalized recommendation~\cite{cui2022m6,bao2023tallrec,liu2023linrec}, customer service~\cite{lu2020improving}, and online advertising~\cite{muhamed2021ctr,yan2025unlocking}. For example, LLMs may act as an intelligent shopping guide to provide specialized and objective recommendations to users in a specific e-commerce industry.
Apart from the advantages brought by LLMs, the factuality issue, e.g., hallucination, has arisen~\cite{wang2023survey,zhao2023felm}. LLMs may provide factually incorrect, unverified, or non-existent information to the users, which may mislead users and cause a catastrophic impact on user experience and merchant and platform revenue. 
Some recent works make efforts to evaluate the factuality of LLMs. 
Previously, human evaluations have been leveraged as a reliable approach to reflect the factuality of LLMs~\cite{yu2022generate,liu2023evaluating,liu2025uqabench}. 
Then, some works try to facilitate the evaluation process by constructing benchmarks using factual knowledge from the open domain, with the aid of retrieved references or other LLMs~\cite{deng2023knowledge,liu2022multi,glover2022revisiting,lee2022factuality,chen2023beyond}. 

Nonetheless, the aforementioned methods face three main challenges in effectively assessing the capabilities of LLMs in e-commerce. Firstly, there is a \textbf{lack of reliability} in both question construction and answer evaluation, which may lead to fluctuating assessments and is not conducive to horizontal comparisons of different models~\cite{chen2023beyond, liu2024evaluating}. For example, some proposed questions may lack ground truth, and the generated ``answer'' from retrieval models, such as LLMs, may introduce unreliability. Additionally, questions that are either too difficult or too simple are ineffective in distinguishing model capabilities. Secondly, \textbf{high consumption} poses challenges in scaling up and accelerating the evaluation. Specifically, existing methods~\cite{liu2023evaluating, lee2022factuality, chen2023beyond, liu2024evaluating} require introducing LLM-as-judge or Human-as-judge to ensure evaluation reliability, given that the responses generated by LLMs are unrestricted. Thirdly, a \textbf{lack of domain expertise} hampers specialized evaluation and optimization in the e-commerce domain. The issue of hallucination in LLMs is particularly urgent in e-commerce~\cite{ding2024intentionqa, jin2024shopping, chen2025chineseecomqa}, directly affecting user experience, merchant revenue, and platform reputation, yet there is a lack of specialized evaluation methods to address this.


In light of the above motivations, we propose a novel benchmark, ECKGBench, to bridge the evaluation gap. We identify a trade-off between reliability and efficiency in existing evaluation paradigms. Without reliable knowledge sources, the evaluation falls into a dilemma: either retrieving unreliable and biased references to ensure efficiency or introducing LLM-as-judge/Human-as-judge to ensure reliability but bringing high evaluation consumption and additional bias. To this end, we employ a large-scale knowledge graph as the foundation of the evaluation and adopt a multiple-choice paradigm for question construction, thus addressing this dilemma at its root. Specifically, the relations in the knowledge graph enable us to automatically generate questions and acquire ground truths without requiring reference generation from retrieval models, guaranteeing evaluation reliability. Additionally, multiple-choice questions provide candidate answers for LLMs, avoiding unconstrained generation that requires an additional judging process to improve efficiency. To further enhance reliability, we improve question quality at multiple stages, including meticulous relation templating, negative sampling, and prompt design. We also incorporate domain expertise in conducting each aforementioned stage. Moreover, we apply domain expertise to categorize questions into common and abstract e-commerce knowledge, allowing for the assessment of LLMs across different dimensions. Besides addressing the above issues, we provide practical methods and metrics to explore the knowledge boundary of the base models of LLMs in e-commerce.


The contributions to our work are four-fold:
\begin{itemize}[leftmargin=*]
\item We propose a benchmark dataset, ECKGBench, which uses a knowledge graph to explore LLMs capabilities in e-commerce. It is the inaugural Chinese benchmark dataset supporting factuality evaluation on e-commerce, whose question reflects real user preference and product popularity.  
\item We propose an automated framework for generating high-quality questions from KG triples. The framework comprises a question generation workflow and a three-stage negative sampling workflow. 
\item We design criteria to detect the base models' knowledge boundary, which is conducive to locating the current model capability and determining the optimization direction.
This promotes the development and evaluation of base models of LLMs.  
\item We comprehensively evaluate LLMs' capability in the e-commerce domain, where we conduct distinctive experiments for fine-tuned and base models. Moreover, we conduct ablation and efficiency studies to demonstrate the efficiency and effectiveness of our proposed ECKGBench. 
\end{itemize}

\section{ECKGBench}

\begin{table}
    \caption{\label{tab:benchmark}A comparison of factuality benchmarks from three perspectives: whether leveraging knowledge graph, specializing in e-commerce, and conducting fine negative sampling.}
    \centering
    \resizebox{0.475\textwidth}{!}{%
    \begin{tabular}{cccc}
        \toprule
        Benchmark & KG-based & E-commerce & Fine-sampled \\
        \cmidrule(lr){1-4}
        FEVER~\cite{thorne2018fever}     & \xmark & \xmark  & \xmark   \\
        WICE~\cite{kamoi2023wice}       & \xmark & \xmark  & \xmark   \\
        HaluEval~\cite{li2023halueval}    & \xmark & \xmark  & \xmark   \\
        FELM~\cite{zhao2024felm}      & \xmark & \xmark  & \xmark   \\
        \cmidrule(lr){1-4}
        GraphEval~\cite{liu2024evaluating}      & \cmark & \xmark  & \xmark   \\
        Head-to-Tail~\cite{sun2023head}   & \cmark & \xmark  & \xmark   \\
        KaRR~\cite{dong2024statistical}       & \cmark & \xmark  & \xmark   \\
        IntentionQA~\cite{ding2024intentionqa} & \xmark & \cmark  & \xmark   \\
        ShoppingMMLU~\cite{jin2024shopping} & \xmark & \cmark  & \xmark   \\
        \cmidrule(lr){1-4}
        \textbf{ECKGBench}   & \cmark & \cmark  & \cmark   \\
         \bottomrule
    \end{tabular}
    }
\end{table}

\begin{figure*}[t]
    \centering
    \includegraphics[width=0.95\linewidth]{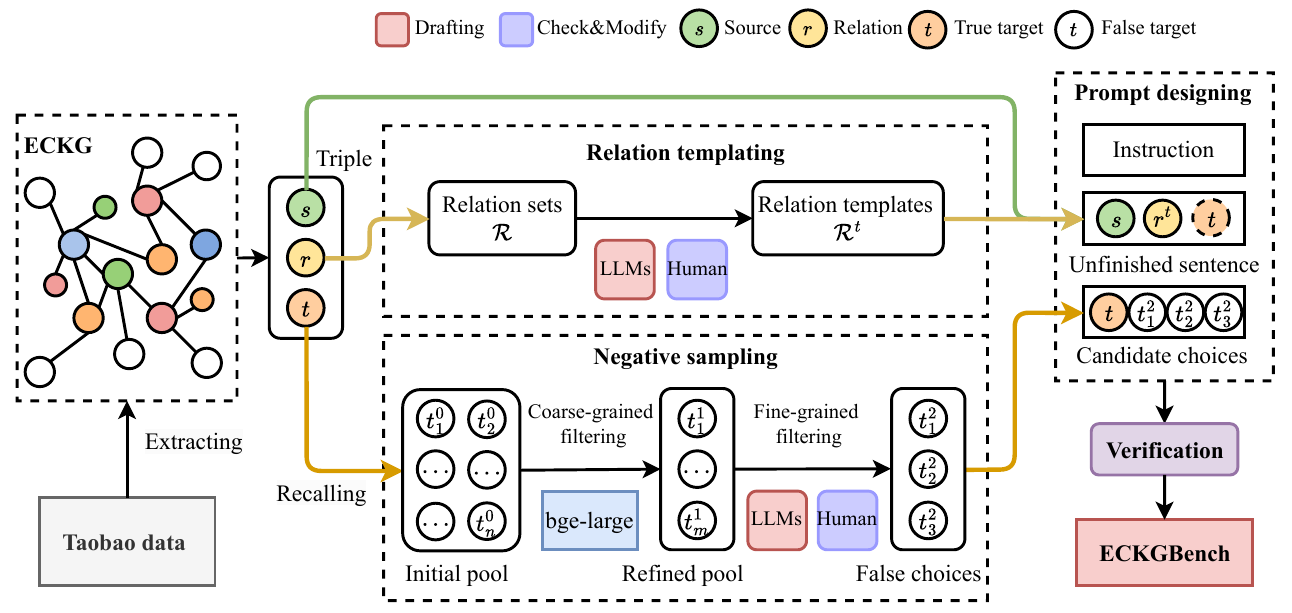}
    \caption{
    An overview of generating the ECKGBench dataset. The sampled triple are fed into question generation (branch above) and negative sampling (branch below) workflows and finally combined to form the questions of ECKGBench.
    }
    \label{fig:overview}
\end{figure*}

In this section, we introduce the proposed ECKGBench dataset, and Figure~\ref{fig:overview} shows its generation overview. First, we provide details of the evaluation source, an e-commerce knowledge graph. 
We will generate interactive question answering based on this KG to assess LLMs.
Specifically, our generation process includes generating relation templates, negative sampling for candidate answers, and assembling accessories to form the final questions. Notably, we propose an automated and hierarchical pipeline for negative sampling, leading to existing factuality benchmarks (the comparison of evaluation methods is demonstrated in Table~\ref{tab:benchmark}).


\subsection{Data Source: E-commerce Knowledge Graph}

The large-scale e-commerce knowledge graph, ECKG\footnote{We will release it in the future to facilitate this topic.}, is collected and extracted from Taobao, containing e-commerce knowledge and concepts, aiming for a lay foundation for improving search accuracy. 
We briefly introduce the development of ECKG. First, we define the types of entities in Taobao, such as item category, function, style, design, color, and user group (e.g., children, adults, elderly). Then, we design the relation types according to the entities' potential links. Based on the pre-defined entities and relations, the system automatically generates $4.8M$ candidate KG triples, formed as $\langle$\textit{source entity, relation, target entity}$\rangle$. The annotator manually annotates the candidate triple with "true" and "false" to indicate whether this is a true triple.

Statistically, ECKG consists of 4.8M candidate KG triples, and 2.1M are true triples. For example, a true triple is like $\langle$\textit{leather case, need, heat insulation}$\rangle$ or $\langle$\textit{cloud white, similar, milk white}$\rangle$.
There are $15$ relations that depict the link between entities and can be divided into two categories: (1) common relations occurring in user living consumption scenarios, such as \textit{(category) has (function)}, \textit{(style) suitable for (group)}, \textit{(group) need (category)}. (2) Abstract relations embody the connections between e-commerce concepts, such as \textit{(color) belonging (color)} and \textit{(style) similar to (style)}. 
Apart from the triple itself, each has additional descriptions (e.g., related item category), improving clarity.

\subsection{Relation Templating}
To evaluate LLM's capabilities, we need to convert the KG triples into a form that aligns with LLM's parametrized nature, i.e., natural language. The parts of a triple, source entity, relation, and target entity correspond to the subject, predicate, and object in a declarative sentence, respectively. 
Nevertheless, the relations $\mathcal{R} = \{r_1,\cdots,r_{\vert\mathcal{R}\vert}\}$ stored in the ECKG are highly condensed, which are inappropriate to directly role as the predicate. Consequently, we employ several LLMs, including GPT-4, Claude 3.5, Qwen2.5-Max, and DeepSeek-R1, to draft multiple candidate templates for each relation. Afterward, we manually select the best candidate and modify the chosen template if necessary. For example, the relation \textit{(function) need (category)} describes that some category of items needs to be equipped with some function, so we select ``\textit{is important to or needed to}'' as the template. This way, we obtain all relation templates, denoted as $\mathcal{R}^t$.

\subsection{Negative Sampling}
Though templates and expert knowledge can determine other parts of prompts, one core challenge to be solved is negative sampling, which is decisive for the quality of the questions. Poor-quality prompts can lead to inconsistent results of LLMs, jeopardizing evaluation reliability. Therefore, we propose a three-stage automated negative sampling workflow to improve the quality of the questions.  

\subsubsection{\textbf{Stage 1: Sample Pool Generating}}
The first challenge is constructing the sampling pool, which lies at the foundation of negative sampling. Clustering-like methods that find the neighbors of ground truth as the sampling pool are unsuitable. First, these methods cannot guarantee that the generated neighbor words are in the e-commerce domain. Second, there is no guarantee that such methods can find negative neighbors since questions may possess multiple answers. Fortunately, we can use the existing knowledge in e-commerce KG. Recall triples of e-commerce KG are labeled as related (i.e., true) and unrelated (false). Then, sampling a true KG triple $\langle s,r,t_0\rangle$, we can use its source entity $s$ and relation $r$ to recall many target entities $\{t_1,\cdots,t_n\}$ from other false KG triples, forming the negative sampling pool, where $n$ is the size. 

\subsubsection{\textbf{Stage 2: Coarse-grained Filtering}}
The question difficulty will fluctuate using random sampling to generate negative samples. For example, the sentence ``\textit{flame red (color) belongs to or is similar to \underline{\hspace{1cm}} (color)}'' with ground truth \textit{rhyme red}. Random sampling may give false choices like \textit{dawn red, dusky red, blush red}, and false choices like \textit{blue, yellow, green}. The former is too hard, while the latter is too easy, and both are inconducive to evaluation reliability. 
Our solution is to filter out extremely hard or easy samples, deeming the semantic similarity between ground truth and negative samples as the measurement of difficulty. First, we choose a suitable embedding model, e.g., bge-large~\cite{xiao2023c}, to obtain the embeddings (with $D$ dimension) of all target entities (including ground truth and negative samples). 
Subsequently, we calculate the cosine similarity between the embeddings of ground truth and negative samples as the semantic similarity. 
Then, we use semantic similarity as a filter to coarsely obtain stage-two negative samples $\{t^1_1,\cdots,t^1_m\}$, where $m$ is the number of negative samples after coarse-grained filtering.

\subsubsection{\textbf{Stage 3: Fine-grained Filtering}}
Coarse-grained filtering narrows down the negative samples. However, we need a more meticulous process to generate final high-quality negative samples. The samples obtained by stage two are semantically close enough that it is hard and time-consuming for human beings to distinguish the quality. 
Therefore, we ask LLMs to assist us in selecting negative samples, sufficiently using their vocabulary understanding. In this stage, we use several LLMs, including GPT-4, Claude 3.5, Qwen2.5-Max, and DeepSeek-R1 to select false choices among negative samples $\{t^1_1,\cdots,t^1_m\}$ generated by the previous stage. 
Furthermore, merely asking GPT-4 to select words cannot align with our targets to improve the question quality. Consequently, we give the question as the context and then add requirements as:
\begin{itemize}[leftmargin=*]
\item The negative examples should be inappropriate for completing the sentence in the *question* but possess effective confusion value that makes them hard to distinguish.
\item The three negative examples should demonstrate diversity and cover multiple domains.
\end{itemize}
An example of prompts we leveraged for LLMs to assist in filtering negative candidates is demonstrated in Appendix~\ref{negtive_sampling}. 
Afterward, we obtain the final version of false choices, $\{t^2_1,t^2_2,t^2_3\}$(defaulted to three). For each true KG triple, the false choices $\{t^2_1,t^2_2,t^2_3\}$ and ground truth $t_0$ are merged to form the corresponding question's multiple choices.

\subsection{Prompt designing}
We are ready to design the prompt and assemble the questions.

\subsubsection{\textbf{Paradigm}}
Letting LLM generate without restriction requires additional assessment from humans or a judge model (e.g., another LLM), leading to excess cost and jeopardizing objectivity.
We apply a multiple-choice form to address this issue, providing better differentiation of LLMs' capabilities compared to true or false questions.

\subsubsection{\textbf{Instruction}}
Instruction is needed to activate the LLMs' knowledge of e-commerce and guide them in answering special questions (e.g., multiple-choice). 
Therefore, we first draft the question instructions manually. Afterward, we leverage several LLMs (e.g., GPT-4, DeepSeek-R1, Claude 3.5) to assist us in refining the details of instructions, such as vocabulary, grammar, and coherence. 

\subsubsection{\textbf{Assembly}}
The prompt's main body is an unfinished declarative sentence consisting of source entity $s$, relation template $r^t$, and the uncompleted part.
The candidate choices consist of the ground truth $t$ and false choices $\{t_1^2,t_2^2,t_3^2\}$. 
The question has a standard multi-choice paradigm with instructions, formed as: [\textit{instruction, sentence, choices}].  
We demonstrate an example in Table~\ref{tab:prompt}.

\begin{table}[ht]
    \caption{An example of assembled question prompt.}
    \label{tab:prompt}
    \centering
    \fontsize{9}{11}\selectfont
    \begin{tabular}{@{}p{8cm}@{}} 
        \toprule
        \textbf{*Instruction*:} Select the most suitable one from among the *Choices* to fill in the blank space to complete the following *Sentence*, making it comply with the logic and knowledge in e-commerce and daily consumption. Just output the selected choice. The terms in brackets are the descriptions of words. \cr
        \textbf{*Sentence*:} Rapid heating (function) is very important for \underline{\hspace{1cm}}  (name of category). \cr
        \textbf{*Options*:} Microwave oven, Flashlight, Electric fan, Speaker \cr
        \textbf{*Answer*:} Microwave oven \cr
        \bottomrule
    \end{tabular}

\end{table}

\subsection{Verification}
The verification process for the evaluation dataset involves a dual-layer approach combining automated LLM analysis and human expert review to ensure robustness and accuracy. 
\subsubsection{\textbf{LLM Verification}} 
Initially, we apply two LLMs, e.g., GPT-4 and DeepSeek, to scan the questions to flag inconsistencies, judging if the answer satisfies the corresponding question. We provide a demonstration of prompting LLMs to assist us in assessing the quality of questions in Appendix~\ref{quality}. We will automatically initially filter out questions regarded as inconsistent by two LLMs simultaneously and record the reasons, and such questions are annotated as low-quality.    
\subsubsection{\textbf{Human Verification}} 
Subsequently, since some questions are related to specialized knowledge in the e-commerce area, human experts will double-check the LLM-identified low-quality questions. The questions that are misjudged as low-quality will be recalled to the question pool. Furthermore, to conform to the regulation, we manually draft the rules to systematically filter out questions that include sensitive words and words unsuitable for children.  

\subsection{Evaluation Dimensions}
\label{sec:dimension}
We evaluate LLMs' capabilities from two e-commerce dimensions: common and abstract knowledge. We provide detailed information on the two dimensions as follows.

\subsubsection{\textbf{Common Knowledge}}
Common knowledge represents users' most cared-for or frequently asked concepts, e.g., a specific product function, a product suitable for which groups, and the suitable season of a category of products. Mastering common knowledge is critical for LLMs in e-commerce since it directly relates to users' first impressions. We collect the corresponding common relations, e.g., \textit{function}, \textit{need}, and \textit{suit}, and then generate questions about common knowledge like Figure~\ref{fig:overview}.    
\subsubsection{\textbf{Abstract Knowledge}}
Abstract knowledge is more complicated and implicit, e.g., one color belongs to another, and one style is similar to another, which is unusual or rare in daily life. Mastering abstract knowledge is the advanced capability to improve implicit revenue, such as recommending products whose colors or styles are similar to what users prefer. We collect abstract relations such as \textit{belong}, \textit{similar}, and \textit{equivalent} and generate questions similarly.  



\begin{table*}
    \caption{\label{tab:know}Types and metrics in exploring knowledge boundary.}
    
    \centering
    \begin{tabular}{lll}
        \toprule
        \textbf{Type\&Metric} & \textbf{Statistical meaning} & \textbf{Semantic meaning} \\
        \cmidrule(lr){1-3}
        WK & $P_{model}(GT(q)\vert q,T=T_0) = 1$ & Always predicts the correct answer. \cr
        SK & $P_{model}(GT(q)\vert q,T=T_0) \in(0,1)$ & Sometimes predicts the correct answer.  \cr
        UK & $P_{model}(GT(q)\vert q,T=T_0) = 0$ & Never predicts the correct answer. \cr
        \cmidrule(lr){1-3}
        SC@k & $\frac{1}{\vert\boldsymbol{q}\vert} \sum_{q_i\in\boldsymbol{q}}\bigwedge_{j=1}^{k} I\big(R^j(q_i)=GT(q_i)\big)$  & All answers are correct in $N$ returns. \cr
        Precision@k & $\frac{1}{\vert\boldsymbol{q}\vert} \sum_{q_i\in\boldsymbol{q}}\sum_{j=1}^k I\big(R^j(q_i)=GT(q_i)\big)/k$ & Rate of correct answers in $N$ returns. \cr
        Recall@k & $\frac{1}{\vert\boldsymbol{q}\vert} \sum_{q_i\in\boldsymbol{q}}{\bigvee_{j=1}^{k} I\big(R^j(q_i)=GT(q_i)\big)}$  & At least one correct answer in $N$ returns. \cr
        \bottomrule
    \end{tabular}

\end{table*}

\section{In-depth Analysis: Knowledge Boundary}
\label{sec:knowledge}
We want to explore the knowledge boundary of LLMs' base models in e-commerce. We first define three non-overlapping knowledge types, based on which we propose three metrics to measure the knowledge boundary.

\subsection{Knowledge Types}
An intuitive way to explore the knowledge boundary is by quantifying the LLMs' mastery level in each knowledge point (i.e., question). Inspired by SliCK~\cite{gekhman2024does}, we divide the knowledge into three categories: \textbf{WK} (well-known), \textbf{SK} (somewhat known), and \textbf{UK} (unknown). To evaluate how well that model knows questions, we set a positive reference temperature $T_0$ to encourage diverse answers. We establish the classification criteria of knowledge types in Table~\ref{tab:know}, with their meanings. $P_{model}(GT(q)\vert q,T=T_0)$ means that given a question $q$ and temperature $T_0$, the probability that model returns the correct answer $GT(q)$. We estimate the probability by using multiple returns from the model. For example, if the model always gives the correct answer, the probability $P_{model}(GT(q)\vert q, T=T_0)$ equals $1$, and we say the model well knows this question.     

\subsection{Evaluation Metrics}
We can calculate each knowledge type's ratio to see an overall distribution. However, the ratio cannot intuitively reflect the knowledge boundary. To quantify the knowledge boundary, we define three metrics as illustrated in Table~\ref{tab:know}: \textbf{SC} (strict correct), \textbf{Precision}, and \textbf{Recall}. SC@k, Precision@k, and Recall@k represent the corresponding values for $k$ returns. 
We can easily find the relation between metrics and knowledge types by formulas: $\text{SC}=\text{WK ratio}$, $\text{Recall}=\text{WK ratio}+\text{SK ratio}=1-\text{UK ratio}$. 
Since SC and Recall are conservative and radical, we add Precision as a more balanced metric. From another perspective, Precision is a continuous metric in terms of each knowledge point, while SC and Recall are discrete metrics. 

\subsection{In-depth Explanation}
The value of SC reflects the essential knowledge boundary of the model, inside which the knowledge is sufficiently trained in the pre-training process. The value of Recall reflects the potential knowledge boundary of the model, and the range between the potential and essential boundary is the knowledge that the model has seen in the pre-training corpus but is unfamiliar. This part of knowledge can be enhanced by post-training methods, such as fine-tuning and RLHF, or activated by in-context learning methods. Knowledge outside the potential boundary does not exist in the pre-training corpus or is hard to learn for the current model size. 
\begin{table*}
  \caption{\label{tab:overall}
    Results of evaluating LLMs on ECKGBench dataset. Each column represents a single evaluation test that leverages one prompting approach (i.e., zero-shot or few-shot) in one e-commerce dimension (i.e., common or abstract). We also give the weighted average score (Avg. ) under zero-shot and few-shot settings and an overall score for each model.  
  }
\centering
    \begin{tabular}{cccccccc}
        \toprule
        \toprule
        \multirow{2}{*}{Models}&
        \multicolumn{3}{|c}{Zero-shot} & \multicolumn{3}{|c|}{Few-shot} & \multirow{2}{*}{Overall Avg.}\cr
        & Common & Abstract & Avg. & Common & Abstract  & Avg. \cr
        \cmidrule(lr){1-8}
        
        ChatGLM3-6B     & 38.24 & 29.61 & 34.27 & 37.14 & 32.43 & 34.98 & 34.63 \cr
        Gemini-1.5-pro  & 45.87 & 45.27 & 45.59 & 40.65 & 31.20 & 36.31 & 40.95 \cr
        Claude3         & 53.83 & 50.00 & 52.09 & 53.04 & 46.80 & 50.18 & 51.14 \cr
        Yi-1.5-6B       & 42.10 & 48.34 & 45.01 & 40.43 & 42.71 & 41.48 & 43.25\cr
        Yi-1.5-34B      & 54.13 & 52.94 & 53.58 & 58.48 & 50.64 & 54.88 & 54.23\cr
        Llama3-8B       & 32.61 & 37.34 & 34.78 & 45.87 & 53.96 & 49.59 & 42.19 \cr
        Llama3-70B      & 57.98 & 46.94 & 52.02 & 51.36 & 59.71 & 55.25 & 53.64 \cr
        Qwen2-7B        & 60.43 & 56.27 & 58.52 & 58.91 & 50.90 & 55.23 & 56.88 \cr
        Qwen2-72B       & 68.48 & 58.97 & 64.12 & \textbf{68.85} & 62.15 & 65.76 & 64.94 \cr
        Qwen2-max       & \textbf{70.65} & 62.66 & \textbf{66.98} & 68.04 & 64.96 & 66.63 & \textbf{66.81} \cr
        GPT-4 turbo     & 62.83 & 60.87 & 61.93 & 66.52 & 65.47 & 66.04 & 63.99 \cr
        GPT-4           & 63.91 & \textbf{67.77} & 65.69 & 67.61 & \textbf{67.26} & \textbf{67.45} & 66.57 \cr
        \bottomrule
        \bottomrule
    \end{tabular}

\end{table*}




\section{Experiments}
We aim to answer the following research questions:
\begin{itemize}[leftmargin=*]
\item \textbf{RQ1}: What is the overall performance of advanced LLMs in e-commerce domain?
\item \textbf{RQ2}: How is the contribution of each component?
\item \textbf{RQ3}: How to explore the knowledge boundary of LLMs?
\item \textbf{RQ4}: How is the efficiency of evaluation in ECKGBench?
\end{itemize}

\subsection{Experimental Setup}

\subsubsection{\textbf{Models}}
In ECKGBench\footnote{\url{https://github.com/ming429778/ECKGBench}.}, we assess the e-commerce capabilities of existing advanced LLMs, including five close-sourced LLMs, i.e., Gemini-1.5-pro, GPT-4 turbo/GPT-4~\citep{OpenAI2023GPT4} and Claude-3, Qwen2-max~\cite{luo2023reasoning}, and seven open-sourced LLMs, i.e., ChatGLM3-6B~\cite{du2021glm}, Yi-1.5-6B/Yi-1.5-34B~\citep{young2024yi}, Llama3-8B/Llama3-70B~\citep{touvron2023llama}, Qwen2-7B/Qwen2-72B~\cite{luo2023reasoning,yang2024qwen2}. Following the default parameters, we apply the API for large-scale generation for close-sourced LLMs. For open-sourced LLMs, we download and test the model from Huggingface\footnote{\url{https://huggingface.co/}}. Notice that we use \textbf{fine-tuned models} (e.g., Chat, Instruct) in RQ1 for fairness. 
In contrast, in RQ3, we use \textbf{base models} to encourage diversified generation, exploring the knowledge boundary of models.


\subsubsection{\textbf{Metrics}}
We evaluated \textit{Accuracy} in RQ1. The accuracy depends on whether the generation of LLMs conforms to the ground truth. 
In RQ2, we assess the \textit{Quality} of datasets using LLM or human assistance. 
Moreover, we define \textit{Inconsistency} to evaluate the reliability: measure whether the model's answers to two questions are inconsistent, where the questions are generated by fixing KG triples and conducting negative sampling twice (i.e., two questions have different false choices).
In RQ3, we use metrics (\textit{SC, Precsion, Recall}) defined in Table~\ref{tab:know}.


\subsubsection{\textbf{Implementation Details}}
To comprehensively evaluate the capabilities of large language models (LLMs), we employ two prompting approaches: zero-shot and few-shot settings in two dimensions of e-commerce knowledge: common knowledge and abstract knowledge. We demonstrate an example of few-shot prompts in Appendix
We create a series of distinctive evaluation settings by combining the prompting approaches with the knowledge dimensions. This allows for a nuanced assessment of each model's performance across various scenarios.

\subsection{Overall Comparison (RQ1)}

We evaluate several advanced LLMs in the ECKGBench and report the overall performances in Table~\ref{tab:overall}. We can draw some interesting observations from the results:
\begin{itemize}[leftmargin=*]
\item In general, LLMs have a low absolute value of accuracy score (e.g., less than $60$) in most cases. The results demonstrate that the capability of LLMs in the e-commerce domain is still unsatisfactory, which is reasonable since e-commerce scenarios are complicated and some specialized concepts in e-commerce barely exist in the pre-training corpus of LLMs. Notably, Qwen2-max and GPT-4 perform well in most cases. 
\item Regarding two knowledge dimensions in e-commerce, LLMs perform significantly better in common knowledge than abstract knowledge, consistent with the previous analysis. In contrast, the prompting methods do not significantly influence the performance, demonstrating the ECKGBench's high-quality questions and good instruction-following capability of fine-tuned models. 
\item Then, we find that the scaling law of LLMs still holds in the e-commerce domain. LLMs with larger scales perform better than ones with more minor scales. We explain that even though some e-commerce concepts are rare in the world corpus, the large-scale LLM can repeat them several times in its pre-training corpus and strictly remember them, leveraging its powerful ability. 
\end{itemize}


\subsection{Ablation Study (RQ2)}
In this section, we evaluate the contribution of each component of question generation. We first give an overall quality assessment. Then, we focus on negative sampling. 
\subsubsection{\textbf{Quality Evaluation}}
In this section, we conduct experiments to compare the qualities of different prompts. We conduct an ablation study with three variants of the proposed question generation framework, including (1) \textbf{\textit{w/o} prompt designing}: without designing question prompts that only ask to complete an unfinished KG triple, (2) \textbf{\textit{w/o} relation template}: without leveraging relation templates to improve readability, (3) \textbf{\textit{w/o} negative sampling}: without the proposed negative sampling workflow (i.e., using random sampling). 
We report the results of original datasets and variants in Table~\ref{tab:ablation}. 
The results show that each component is indispensable, demonstrating the superiority of the proposed ECKGBench. The prompt design contributes the most to the quality, as it helps make questions conform to natural language and guides LLMs to leverage the parameterized knowledge. 

\begin{table}[t]
\caption{\label{tab:ablation}Ablation of question generation.}
    \centering
    \begin{tabular}{ccc}
        \toprule
        Variants & GPT-4 eval & Human eval \\
        \cmidrule(lr){1-3}
        Ours & \textbf{0.6104} & \textbf{0.92} \\
        \textit{w/o} prompt designing & 0.2074 & 0.34 \\
        \textit{w/o} relation templating & 0.3590 & 0.51 \\
        \textit{w/o} negative sampling & 0.5410 & 0.85 \\
        \bottomrule
    \end{tabular}
\end{table}
\subsubsection{\textbf{Inconsistency Evaluation}}
\begin{figure}[t]
    \centering
    \includegraphics[width=0.9\linewidth]{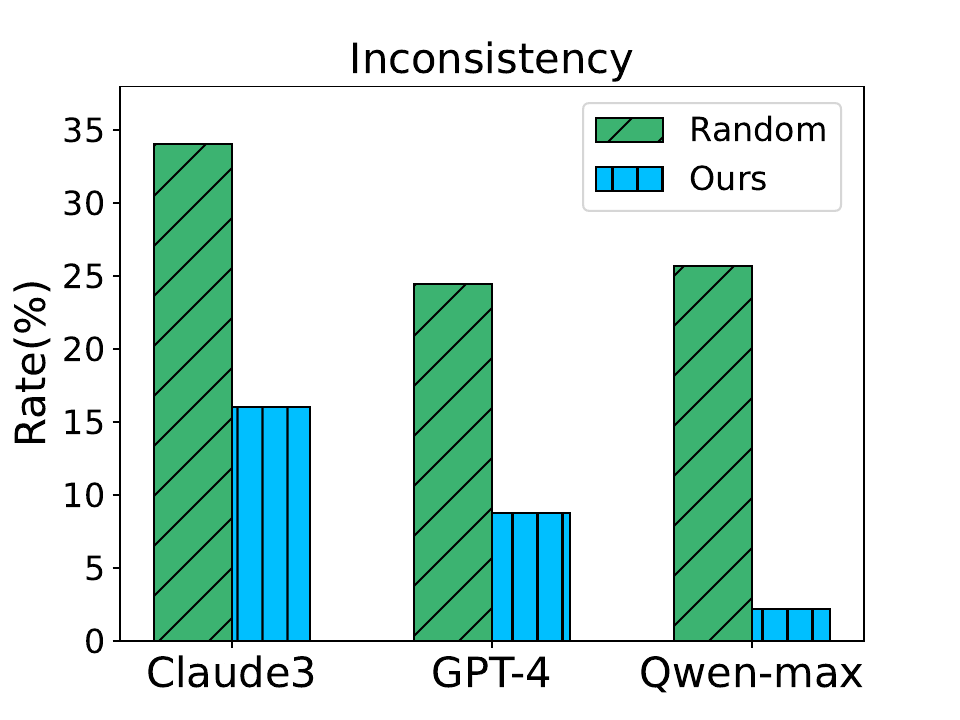}
    \vspace{-3mm}
    \caption{\label{fig:ablation}
    Results of inconsistency rates. The lower, the better. The green and blue represent the random sampling and our sampling methods, respectively.  
    }
\end{figure}

The contribution of sampling workflow to question quality seems relatively minor. It is reasonable since the sampling workflow focuses more on improving stability and reliability. Consequently, we evaluate the LLMs' inconsistency in proposed and random sampling. 
We report the results in Figure~\ref{fig:ablation}. The results demonstrate the effectiveness of our sampling workflow. Compared to random sampling, LLM's answers to the dataset generated by our methods are highly consistent.
Especially, our methods can control the inconsistency rate below $10\%$ for large-scale models (i.e., GPT-4 and Qwen-max). 


\begin{table*}
  \caption{\label{tab:rq3}
     Results of exploring knowledge boundary of Qwen-7B. We report Precision, Recall, and SC in different temperatures. 
  }
\centering
    \begin{tabular}{cccccccccc}
        \toprule
        \toprule
        \multirow{2}{*}{Temperatures}&
        \multicolumn{3}{|c}{{Precision@5}} & \multicolumn{3}{|c}{Recall@5} & \multicolumn{3}{|c}{SC@5}\cr
        & Common & Abstract & Avg. & Common & Abstract & Avg. & Common & Abstract & Avg. \cr
        \cmidrule(lr){1-10}
        $T=0.01$   & 50.22 & 43.58 & 47.17 & 50.65 & 44.25 & 47.71 & 49.78 & 43.22 & 46.77 \cr
        $T=0.1$ & 49.61 & 43.63 & 46.86 & 55.65 & 49.10 & 52.64 & 43.26 & 36.06 & 39.95 \cr
        $T=0.2$ & 50.00 & 42.97 & 46.77 & 61.96 & 53.96 & 58.28 & 38.91 & 30.43 & 35.02 \cr
        $T=0.3$ & 50.26 & 42.25 & 46.58 & 66.30 & 60.36 & 63.57 & 32.39 & 28.64 & 30.67 \cr
        $T=0.5$ & 46.57 & 41.84 & 44.39 & 70.65 & 65.47 & 68.27 & 23.26 & 17.90 & 20.80 \cr
        \bottomrule
        \bottomrule
    \end{tabular}
\end{table*}

\subsection{Exploring Knowledge Boundary (RQ3)}
Then, we explore the knowledge boundary of Qwen-7B. Notice that we use the base model here. In Table~\ref{tab:rq3}, we report the previously defined metrics SC@5, Precision@5, and Recall@5, tuning the model temperature in search range $\{0.01, 0.1, 0.2, 0.3, 0.5\}$.
We observe that due to model generation being encouraged to be diversified by increasing temperatures,  the precision gets slightly reduced, and the Recall and SC (strict correct) increase and reduce significantly. 
Determining reference temperature $T_0$ is empirical, with often a small value such as $0.2$ or $0.3$. 
Here, we set $T_0=0.2$, as in which the Recall is $58.28$, close to the Accuracy $58.52$ of its fine-tuned version (Qwen-7B-Instruct), conforming to our analysis that the Recall reflects the potential knowledge boundary, which post-training methods can reach. We can also know the essential boundary of the model according to SC value $35.02$, inside which the knowledge is sufficiently trained.


\begin{table*}[t]
  \caption{\label{tab:rq4}
     Results of efficiency test. We report the average token number of inputs and outputs and the average response time (Avg. RT). In addition, we record the maximum response time (Max. RT), which is decisive for the time cost of batch API calls. 
  }
\centering
{
    \begin{tabular}{ccccccccc}
        \toprule
        \multirow{2}{*}{Model}&
        \multicolumn{4}{|c}{{Zero-shot}} & \multicolumn{4}{|c}{Few-shot} \cr
        & Input & Output & Avg. RT & Max. RT & Input & Output & Avg. RT & Max. RT \cr
        \cmidrule(lr){1-9}
Gemini    & 91.21  & 87.63 & 2.80  & 5.76 & 225.21 & 45.56 & 2.36  & 6.22\cr
Claude3   & 142.77 & 30.69 & 18.06 & 29.97& 319.26 & 95.40 & 17.83 & 29.51\cr
Llama3-70B& 120.38 & 26.42 & 2.09  & 8.63 & 268.02 & 4.84  & 0.88  & 6.92\cr
Qwen2-72B & 93.55  & 12.24 & 1.05  & 9.72 & 236.55 & 11.00 & 1.76  & 41.77\cr
Qwen2-max & 108.07 & 2.68  & 0.71  & 4.51 & 233.55 & 2.48  & 0.72  & 3.73\cr
GPT-4     & 147.38 & 6.53  & 1.74  & 5.56 & 332.38 & 4.95  & 1.61  & 7.42 \cr
        \cmidrule(lr){1-9}
Avg.      & 117.23 & 27.70 & 4.41 & 10.69 & 269.16 & 27.37 & 4.19 & 16.00 \cr
        \bottomrule
    \end{tabular}}

\end{table*}

\subsection{Efficiency Study (RQ4)}
Efficiency is critical in evaluating LLMs due to the high costs associated with their inference. In this context, we examine the efficiency of the proposed ECKGBench. The results are presented in Table~\ref{tab:rq4}, illustrating the effectiveness of our evaluation framework in reducing cost.
Notably, in most cases, the average response times under zero-shot and few-shot settings are below 3 seconds, with maximum response times capped at under 10 seconds. This rapid assessment significantly enhances the usability of ECKGBench without sacrificing reliability. Moreover, our question refinement effectively controls the input and output token numbers, contributing to overall efficiency. 

\section{Related Work}
\subsection{Factuality Evaluation of LLMs}
With the prevailing of LLMs in NLP tasks, one crucial concern reveals that the LLMs may hallucinate in some scenarios~\citep{wei2022chain,li2023halueval,zhang2023siren,chen2023beyond,dong2024statistical}. Hallucinations are related to LLMs' factuality capabilities. Many efforts dive into evaluation benchmarks or methods to explore the factuality boundary of LLMs. 
Early works propose human evaluation protocols to assess LLMs performance in factuality~\citep{li2022eliciting,yu2022generate,liu2023evaluating}. However, the results of such an approach may fluctuate with the varying question quality and human subjectivity. In addition, this approach is too costly and time-consuming to scale up. 
Another line of work designs the dataset using sources from the open domain~\citep{deng2023knowledge,liu2022multi,glover2022revisiting,lee2022factuality,manakul2023selfcheckgpt}. The designed dataset is helpful for automated evaluation, which boosts scaling. Again, human-labeled references may limit the objectivity and generalization of the evaluation.  
Some works further reduce the role of human assistance by introducing retrieval models or external LLMs, e.g., GPT-4, to generate references automatically~\citep{glover2022revisiting,lee2022factuality,chen2023beyond,liu2024evaluating}. While more objective than human labeling, these methods still depend on the retrieval models or LLMs. 


\subsection{Evaluation of LLMs using KG}
Knowledge graphs (KGs) consist of highly condensed factual knowledge, typically represented as triples~\citep{bordes2013translating,sun2019rotate}. Existing research primarily focuses on leveraging KGs as supplementary knowledge to enhance the factuality performance of LLMs~\citep{yasunaga2022deep,jiang2023reasoninglm,luo2023reasoning}. Another promising direction involves utilizing KGs~\citep{sun2023head,liu2024evaluating,dong2024statistical} to evaluate the factuality of LLMs. However, most existing KG-based benchmarks are geared toward open domains and rely on random negative sampling, possessing a gap in assessing the actual capabilities of LLMs within specific domains, such as e-commerce.


\section{Conclusion}

In the paper, we propose a brand new benchmark for evaluating LLMs in e-commerce, named ECKGBench. We meticulously generate the relation templates, a necessary step for transforming the KG triples
into a form that aligns with LLM's parametrized nature. In addition, to improve question quality, we propose a three-stage negative sampling workflow to automatically generate high-quality false choices, thus improving the reliability and stability of the evaluation. Afterward, we carefully designed the question prompt and conducted thorough verification to guarantee the questions were of high quality. Beyond the design of regular assessment, we propose a novel approach to evaluate base LLM's knowledge boundary in e-commerce.
We hope the proposed ECKGBench will benefit the evaluation of LLMs in the e-commerce domain.


\newpage
\bibliographystyle{ACM-Reference-Format}
\bibliography{custom}


\begin{thebibliography}{44}


\ifx \showCODEN    \undefined \def \showCODEN     #1{\unskip}     \fi
\ifx \showDOI      \undefined \def \showDOI       #1{#1}\fi
\ifx \showISBNx    \undefined \def \showISBNx     #1{\unskip}     \fi
\ifx \showISBNxiii \undefined \def \showISBNxiii  #1{\unskip}     \fi
\ifx \showISSN     \undefined \def \showISSN      #1{\unskip}     \fi
\ifx \showLCCN     \undefined \def \showLCCN      #1{\unskip}     \fi
\ifx \shownote     \undefined \def \shownote      #1{#1}          \fi
\ifx \showarticletitle \undefined \def \showarticletitle #1{#1}   \fi
\ifx \showURL      \undefined \def \showURL       {\relax}        \fi
\providecommand\bibfield[2]{#2}
\providecommand\bibinfo[2]{#2}
\providecommand\natexlab[1]{#1}
\providecommand\showeprint[2][]{arXiv:#2}

\bibitem[Bao et~al\mbox{.}(2023)]%
        {bao2023tallrec}
\bibfield{author}{\bibinfo{person}{Keqin Bao}, \bibinfo{person}{Jizhi Zhang}, \bibinfo{person}{Yang Zhang}, \bibinfo{person}{Wenjie Wang}, \bibinfo{person}{Fuli Feng}, {and} \bibinfo{person}{Xiangnan He}.} \bibinfo{year}{2023}\natexlab{}.
\newblock \showarticletitle{Tallrec: An effective and efficient tuning framework to align large language model with recommendation}. In \bibinfo{booktitle}{\emph{Proceedings of the 17th ACM Conference on Recommender Systems}}. \bibinfo{pages}{1007--1014}.
\newblock


\bibitem[Bordes et~al\mbox{.}(2013)]%
        {bordes2013translating}
\bibfield{author}{\bibinfo{person}{Antoine Bordes}, \bibinfo{person}{Nicolas Usunier}, \bibinfo{person}{Alberto Garcia-Duran}, \bibinfo{person}{Jason Weston}, {and} \bibinfo{person}{Oksana Yakhnenko}.} \bibinfo{year}{2013}\natexlab{}.
\newblock \showarticletitle{Translating embeddings for modeling multi-relational data}.
\newblock \bibinfo{journal}{\emph{Advances in neural information processing systems}}  \bibinfo{volume}{26} (\bibinfo{year}{2013}).
\newblock


\bibitem[Chen et~al\mbox{.}(2025)]%
        {chen2025chineseecomqa}
\bibfield{author}{\bibinfo{person}{Haibin Chen}, \bibinfo{person}{Kangtao Lv}, \bibinfo{person}{Chengwei Hu}, \bibinfo{person}{Yanshi Li}, \bibinfo{person}{Yujin Yuan}, \bibinfo{person}{Yancheng He}, \bibinfo{person}{Xingyao Zhang}, \bibinfo{person}{Langming Liu}, \bibinfo{person}{Shilei Liu}, \bibinfo{person}{Wenbo Su}, {et~al\mbox{.}}} \bibinfo{year}{2025}\natexlab{}.
\newblock \showarticletitle{ChineseEcomQA: A Scalable E-commerce Concept Evaluation Benchmark for Large Language Models}.
\newblock \bibinfo{journal}{\emph{arXiv preprint arXiv:2502.20196}} (\bibinfo{year}{2025}).
\newblock


\bibitem[Chen et~al\mbox{.}(2023)]%
        {chen2023beyond}
\bibfield{author}{\bibinfo{person}{Liang Chen}, \bibinfo{person}{Yang Deng}, \bibinfo{person}{Yatao Bian}, \bibinfo{person}{Zeyu Qin}, \bibinfo{person}{Bingzhe Wu}, \bibinfo{person}{Tat-Seng Chua}, {and} \bibinfo{person}{Kam-Fai Wong}.} \bibinfo{year}{2023}\natexlab{}.
\newblock \showarticletitle{Beyond factuality: A comprehensive evaluation of large language models as knowledge generators}.
\newblock \bibinfo{journal}{\emph{arXiv preprint arXiv:2310.07289}} (\bibinfo{year}{2023}).
\newblock


\bibitem[Chowdhery et~al\mbox{.}(2023)]%
        {chowdhery2023palm}
\bibfield{author}{\bibinfo{person}{Aakanksha Chowdhery}, \bibinfo{person}{Sharan Narang}, \bibinfo{person}{Jacob Devlin}, \bibinfo{person}{Maarten Bosma}, \bibinfo{person}{Gaurav Mishra}, \bibinfo{person}{Adam Roberts}, \bibinfo{person}{Paul Barham}, \bibinfo{person}{Hyung~Won Chung}, \bibinfo{person}{Charles Sutton}, \bibinfo{person}{Sebastian Gehrmann}, {et~al\mbox{.}}} \bibinfo{year}{2023}\natexlab{}.
\newblock \showarticletitle{Palm: Scaling language modeling with pathways}.
\newblock \bibinfo{journal}{\emph{Journal of Machine Learning Research}} \bibinfo{volume}{24}, \bibinfo{number}{240} (\bibinfo{year}{2023}), \bibinfo{pages}{1--113}.
\newblock


\bibitem[Cui et~al\mbox{.}(2022)]%
        {cui2022m6}
\bibfield{author}{\bibinfo{person}{Zeyu Cui}, \bibinfo{person}{Jianxin Ma}, \bibinfo{person}{Chang Zhou}, \bibinfo{person}{Jingren Zhou}, {and} \bibinfo{person}{Hongxia Yang}.} \bibinfo{year}{2022}\natexlab{}.
\newblock \showarticletitle{M6-rec: Generative pretrained language models are open-ended recommender systems}.
\newblock \bibinfo{journal}{\emph{arXiv preprint arXiv:2205.08084}} (\bibinfo{year}{2022}).
\newblock


\bibitem[Deng et~al\mbox{.}(2023)]%
        {deng2023knowledge}
\bibfield{author}{\bibinfo{person}{Yang Deng}, \bibinfo{person}{Wenxuan Zhang}, \bibinfo{person}{Yifei Yuan}, {and} \bibinfo{person}{Wai Lam}.} \bibinfo{year}{2023}\natexlab{}.
\newblock \showarticletitle{Knowledge-enhanced mixed-initiative dialogue system for emotional support conversations}.
\newblock \bibinfo{journal}{\emph{arXiv preprint arXiv:2305.10172}} (\bibinfo{year}{2023}).
\newblock


\bibitem[Ding et~al\mbox{.}(2024)]%
        {ding2024intentionqa}
\bibfield{author}{\bibinfo{person}{Wenxuan Ding}, \bibinfo{person}{Weiqi Wang}, \bibinfo{person}{Sze Heng~Douglas Kwok}, \bibinfo{person}{Minghao Liu}, \bibinfo{person}{Tianqing Fang}, \bibinfo{person}{Jiaxin Bai}, \bibinfo{person}{Junxian He}, {and} \bibinfo{person}{Yangqiu Song}.} \bibinfo{year}{2024}\natexlab{}.
\newblock \showarticletitle{IntentionQA: A Benchmark for Evaluating Purchase Intention Comprehension Abilities of Language Models in E-commerce}.
\newblock \bibinfo{journal}{\emph{arXiv preprint arXiv:2406.10173}} (\bibinfo{year}{2024}).
\newblock


\bibitem[Dong et~al\mbox{.}(2024)]%
        {dong2024statistical}
\bibfield{author}{\bibinfo{person}{Qingxiu Dong}, \bibinfo{person}{Jingjing Xu}, \bibinfo{person}{Lingpeng Kong}, \bibinfo{person}{Zhifang Sui}, {and} \bibinfo{person}{Lei Li}.} \bibinfo{year}{2024}\natexlab{}.
\newblock \showarticletitle{Statistical Knowledge Assessment for Large Language Models}.
\newblock \bibinfo{journal}{\emph{Advances in Neural Information Processing Systems}}  \bibinfo{volume}{36} (\bibinfo{year}{2024}).
\newblock


\bibitem[Du et~al\mbox{.}(2021)]%
        {du2021glm}
\bibfield{author}{\bibinfo{person}{Zhengxiao Du}, \bibinfo{person}{Yujie Qian}, \bibinfo{person}{Xiao Liu}, \bibinfo{person}{Ming Ding}, \bibinfo{person}{Jiezhong Qiu}, \bibinfo{person}{Zhilin Yang}, {and} \bibinfo{person}{Jie Tang}.} \bibinfo{year}{2021}\natexlab{}.
\newblock \showarticletitle{Glm: General language model pretraining with autoregressive blank infilling}.
\newblock \bibinfo{journal}{\emph{arXiv preprint arXiv:2103.10360}} (\bibinfo{year}{2021}).
\newblock


\bibitem[Gekhman et~al\mbox{.}(2024)]%
        {gekhman2024does}
\bibfield{author}{\bibinfo{person}{Zorik Gekhman}, \bibinfo{person}{Gal Yona}, \bibinfo{person}{Roee Aharoni}, \bibinfo{person}{Matan Eyal}, \bibinfo{person}{Amir Feder}, \bibinfo{person}{Roi Reichart}, {and} \bibinfo{person}{Jonathan Herzig}.} \bibinfo{year}{2024}\natexlab{}.
\newblock \showarticletitle{Does Fine-Tuning LLMs on New Knowledge Encourage Hallucinations?}
\newblock \bibinfo{journal}{\emph{arXiv preprint arXiv:2405.05904}} (\bibinfo{year}{2024}).
\newblock


\bibitem[Glover et~al\mbox{.}(2022)]%
        {glover2022revisiting}
\bibfield{author}{\bibinfo{person}{John Glover}, \bibinfo{person}{Federico Fancellu}, \bibinfo{person}{Vasudevan Jagannathan}, \bibinfo{person}{Matthew~R Gormley}, {and} \bibinfo{person}{Thomas Schaaf}.} \bibinfo{year}{2022}\natexlab{}.
\newblock \showarticletitle{Revisiting text decomposition methods for NLI-based factuality scoring of summaries}.
\newblock \bibinfo{journal}{\emph{arXiv preprint arXiv:2211.16853}} (\bibinfo{year}{2022}).
\newblock


\bibitem[Guu et~al\mbox{.}(2020)]%
        {guu2020retrieval}
\bibfield{author}{\bibinfo{person}{Kelvin Guu}, \bibinfo{person}{Kenton Lee}, \bibinfo{person}{Zora Tung}, \bibinfo{person}{Panupong Pasupat}, {and} \bibinfo{person}{Mingwei Chang}.} \bibinfo{year}{2020}\natexlab{}.
\newblock \showarticletitle{Retrieval augmented language model pre-training}. In \bibinfo{booktitle}{\emph{International conference on machine learning}}. PMLR, \bibinfo{pages}{3929--3938}.
\newblock


\bibitem[Jiang et~al\mbox{.}(2023)]%
        {jiang2023reasoninglm}
\bibfield{author}{\bibinfo{person}{Jinhao Jiang}, \bibinfo{person}{Kun Zhou}, \bibinfo{person}{Wayne~Xin Zhao}, \bibinfo{person}{Yaliang Li}, {and} \bibinfo{person}{Ji-Rong Wen}.} \bibinfo{year}{2023}\natexlab{}.
\newblock \showarticletitle{ReasoningLM: Enabling Structural Subgraph Reasoning in Pre-trained Language Models for Question Answering over Knowledge Graph}.
\newblock \bibinfo{journal}{\emph{arXiv preprint arXiv:2401.00158}} (\bibinfo{year}{2023}).
\newblock


\bibitem[Jin et~al\mbox{.}(2024)]%
        {jin2024shopping}
\bibfield{author}{\bibinfo{person}{Yilun Jin}, \bibinfo{person}{Zheng Li}, \bibinfo{person}{Chenwei Zhang}, \bibinfo{person}{Tianyu Cao}, \bibinfo{person}{Yifan Gao}, \bibinfo{person}{Pratik Jayarao}, \bibinfo{person}{Mao Li}, \bibinfo{person}{Xin Liu}, \bibinfo{person}{Ritesh Sarkhel}, \bibinfo{person}{Xianfeng Tang}, {et~al\mbox{.}}} \bibinfo{year}{2024}\natexlab{}.
\newblock \showarticletitle{Shopping MMLU: A Massive Multi-Task Online Shopping Benchmark for Large Language Models}.
\newblock \bibinfo{journal}{\emph{arXiv preprint arXiv:2410.20745}} (\bibinfo{year}{2024}).
\newblock


\bibitem[Kamoi et~al\mbox{.}(2023)]%
        {kamoi2023wice}
\bibfield{author}{\bibinfo{person}{Ryo Kamoi}, \bibinfo{person}{Tanya Goyal}, \bibinfo{person}{Juan~Diego Rodriguez}, {and} \bibinfo{person}{Greg Durrett}.} \bibinfo{year}{2023}\natexlab{}.
\newblock \showarticletitle{Wice: Real-world entailment for claims in wikipedia}.
\newblock \bibinfo{journal}{\emph{arXiv preprint arXiv:2303.01432}} (\bibinfo{year}{2023}).
\newblock


\bibitem[Lee et~al\mbox{.}(2022)]%
        {lee2022factuality}
\bibfield{author}{\bibinfo{person}{Nayeon Lee}, \bibinfo{person}{Wei Ping}, \bibinfo{person}{Peng Xu}, \bibinfo{person}{Mostofa Patwary}, \bibinfo{person}{Pascale~N Fung}, \bibinfo{person}{Mohammad Shoeybi}, {and} \bibinfo{person}{Bryan Catanzaro}.} \bibinfo{year}{2022}\natexlab{}.
\newblock \showarticletitle{Factuality enhanced language models for open-ended text generation}.
\newblock \bibinfo{journal}{\emph{Advances in Neural Information Processing Systems}}  \bibinfo{volume}{35} (\bibinfo{year}{2022}), \bibinfo{pages}{34586--34599}.
\newblock


\bibitem[Li et~al\mbox{.}(2023)]%
        {li2023halueval}
\bibfield{author}{\bibinfo{person}{Junyi Li}, \bibinfo{person}{Xiaoxue Cheng}, \bibinfo{person}{Wayne~Xin Zhao}, \bibinfo{person}{Jian-Yun Nie}, {and} \bibinfo{person}{Ji-Rong Wen}.} \bibinfo{year}{2023}\natexlab{}.
\newblock \showarticletitle{Halueval: A large-scale hallucination evaluation benchmark for large language models}. In \bibinfo{booktitle}{\emph{Proceedings of the 2023 Conference on Empirical Methods in Natural Language Processing}}. \bibinfo{pages}{6449--6464}.
\newblock


\bibitem[Li et~al\mbox{.}(2022)]%
        {li2022eliciting}
\bibfield{author}{\bibinfo{person}{Yanyang Li}, \bibinfo{person}{Jianqiao Zhao}, \bibinfo{person}{Michael~R Lyu}, {and} \bibinfo{person}{Liwei Wang}.} \bibinfo{year}{2022}\natexlab{}.
\newblock \showarticletitle{Eliciting knowledge from large pre-trained models for unsupervised knowledge-grounded conversation}.
\newblock \bibinfo{journal}{\emph{arXiv preprint arXiv:2211.01587}} (\bibinfo{year}{2022}).
\newblock


\bibitem[Liu et~al\mbox{.}(2023a)]%
        {liu2023linrec}
\bibfield{author}{\bibinfo{person}{Langming Liu}, \bibinfo{person}{Liu Cai}, \bibinfo{person}{Chi Zhang}, \bibinfo{person}{Xiangyu Zhao}, \bibinfo{person}{Jingtong Gao}, \bibinfo{person}{Wanyu Wang}, \bibinfo{person}{Yifu Lv}, \bibinfo{person}{Wenqi Fan}, \bibinfo{person}{Yiqi Wang}, \bibinfo{person}{Ming He}, {et~al\mbox{.}}} \bibinfo{year}{2023}\natexlab{a}.
\newblock \showarticletitle{Linrec: Linear attention mechanism for long-term sequential recommender systems}. In \bibinfo{booktitle}{\emph{Proceedings of the 46th International ACM SIGIR Conference on Research and Development in Information Retrieval}}. \bibinfo{pages}{289--299}.
\newblock


\bibitem[Liu et~al\mbox{.}(2025)]%
        {liu2025uqabench}
\bibfield{author}{\bibinfo{person}{Langming Liu}, \bibinfo{person}{Shilei Liu}, \bibinfo{person}{Yujin Yuan}, \bibinfo{person}{Yizhen Zhang}, \bibinfo{person}{Bencheng Yan}, \bibinfo{person}{Zhiyuan Zeng}, \bibinfo{person}{Zihao Wang}, \bibinfo{person}{Jiaqi Liu}, \bibinfo{person}{Di Wang}, \bibinfo{person}{Wenbo Su}, {et~al\mbox{.}}} \bibinfo{year}{2025}\natexlab{}.
\newblock \showarticletitle{UQABench: Evaluating User Embedding for Prompting LLMs in Personalized Question Answering}.
\newblock \bibinfo{journal}{\emph{arXiv preprint arXiv:2502.19178}} (\bibinfo{year}{2025}).
\newblock


\bibitem[Liu et~al\mbox{.}(2023b)]%
        {liu2023evaluating}
\bibfield{author}{\bibinfo{person}{Nelson~F Liu}, \bibinfo{person}{Tianyi Zhang}, {and} \bibinfo{person}{Percy Liang}.} \bibinfo{year}{2023}\natexlab{b}.
\newblock \showarticletitle{Evaluating verifiability in generative search engines}.
\newblock \bibinfo{journal}{\emph{arXiv preprint arXiv:2304.09848}} (\bibinfo{year}{2023}).
\newblock


\bibitem[Liu et~al\mbox{.}(2024)]%
        {liu2024evaluating}
\bibfield{author}{\bibinfo{person}{Xiaoze Liu}, \bibinfo{person}{Feijie Wu}, \bibinfo{person}{Tianyang Xu}, \bibinfo{person}{Zhuo Chen}, \bibinfo{person}{Yichi Zhang}, \bibinfo{person}{Xiaoqian Wang}, {and} \bibinfo{person}{Jing Gao}.} \bibinfo{year}{2024}\natexlab{}.
\newblock \showarticletitle{Evaluating the Factuality of Large Language Models using Large-Scale Knowledge Graphs}.
\newblock \bibinfo{journal}{\emph{arXiv preprint arXiv:2404.00942}} (\bibinfo{year}{2024}).
\newblock


\bibitem[Liu et~al\mbox{.}(2022)]%
        {liu2022multi}
\bibfield{author}{\bibinfo{person}{Zihan Liu}, \bibinfo{person}{Mostofa Patwary}, \bibinfo{person}{Ryan Prenger}, \bibinfo{person}{Shrimai Prabhumoye}, \bibinfo{person}{Wei Ping}, \bibinfo{person}{Mohammad Shoeybi}, {and} \bibinfo{person}{Bryan Catanzaro}.} \bibinfo{year}{2022}\natexlab{}.
\newblock \showarticletitle{Multi-stage prompting for knowledgeable dialogue generation}.
\newblock \bibinfo{journal}{\emph{arXiv preprint arXiv:2203.08745}} (\bibinfo{year}{2022}).
\newblock


\bibitem[Lu et~al\mbox{.}(2020)]%
        {lu2020improving}
\bibfield{author}{\bibinfo{person}{Junyu Lu}, \bibinfo{person}{Xiancong Ren}, \bibinfo{person}{Yazhou Ren}, \bibinfo{person}{Ao Liu}, {and} \bibinfo{person}{Zenglin Xu}.} \bibinfo{year}{2020}\natexlab{}.
\newblock \showarticletitle{Improving contextual language models for response retrieval in multi-turn conversation}. In \bibinfo{booktitle}{\emph{Proceedings of the 43rd International ACM SIGIR Conference on Research and Development in Information Retrieval}}. \bibinfo{pages}{1805--1808}.
\newblock


\bibitem[Luo et~al\mbox{.}(2023)]%
        {luo2023reasoning}
\bibfield{author}{\bibinfo{person}{Linhao Luo}, \bibinfo{person}{Yuan-Fang Li}, \bibinfo{person}{Gholamreza Haffari}, {and} \bibinfo{person}{Shirui Pan}.} \bibinfo{year}{2023}\natexlab{}.
\newblock \showarticletitle{Reasoning on graphs: Faithful and interpretable large language model reasoning}.
\newblock \bibinfo{journal}{\emph{arXiv preprint arXiv:2310.01061}} (\bibinfo{year}{2023}).
\newblock


\bibitem[Manakul et~al\mbox{.}(2023)]%
        {manakul2023selfcheckgpt}
\bibfield{author}{\bibinfo{person}{Potsawee Manakul}, \bibinfo{person}{Adian Liusie}, {and} \bibinfo{person}{Mark~JF Gales}.} \bibinfo{year}{2023}\natexlab{}.
\newblock \showarticletitle{Selfcheckgpt: Zero-resource black-box hallucination detection for generative large language models}.
\newblock \bibinfo{journal}{\emph{arXiv preprint arXiv:2303.08896}} (\bibinfo{year}{2023}).
\newblock


\bibitem[Muhamed et~al\mbox{.}(2021)]%
        {muhamed2021ctr}
\bibfield{author}{\bibinfo{person}{Aashiq Muhamed}, \bibinfo{person}{Iman Keivanloo}, \bibinfo{person}{Sujan Perera}, \bibinfo{person}{James Mracek}, \bibinfo{person}{Yi Xu}, \bibinfo{person}{Qingjun Cui}, \bibinfo{person}{Santosh Rajagopalan}, \bibinfo{person}{Belinda Zeng}, {and} \bibinfo{person}{Trishul Chilimbi}.} \bibinfo{year}{2021}\natexlab{}.
\newblock \showarticletitle{CTR-BERT: Cost-effective knowledge distillation for billion-parameter teacher models}. In \bibinfo{booktitle}{\emph{NeurIPS Efficient Natural Language and Speech Processing Workshop}}.
\newblock


\bibitem[OpenAI(2023)]%
        {OpenAI2023GPT4}
\bibfield{author}{\bibinfo{person}{OpenAI}.} \bibinfo{year}{2023}\natexlab{}.
\newblock \showarticletitle{GPT-4 technical report}.
\newblock  (\bibinfo{year}{2023}).
\newblock


\bibitem[Sun et~al\mbox{.}(2023)]%
        {sun2023head}
\bibfield{author}{\bibinfo{person}{Kai Sun}, \bibinfo{person}{Yifan~Ethan Xu}, \bibinfo{person}{Hanwen Zha}, \bibinfo{person}{Yue Liu}, {and} \bibinfo{person}{Xin~Luna Dong}.} \bibinfo{year}{2023}\natexlab{}.
\newblock \showarticletitle{Head-to-tail: How knowledgeable are large language models (llm)? AKA will llms replace knowledge graphs?}
\newblock \bibinfo{journal}{\emph{arXiv preprint arXiv:2308.10168}} (\bibinfo{year}{2023}).
\newblock


\bibitem[Sun et~al\mbox{.}(2019)]%
        {sun2019rotate}
\bibfield{author}{\bibinfo{person}{Zhiqing Sun}, \bibinfo{person}{Zhi-Hong Deng}, \bibinfo{person}{Jian-Yun Nie}, {and} \bibinfo{person}{Jian Tang}.} \bibinfo{year}{2019}\natexlab{}.
\newblock \showarticletitle{Rotate: Knowledge graph embedding by relational rotation in complex space}.
\newblock \bibinfo{journal}{\emph{arXiv preprint arXiv:1902.10197}} (\bibinfo{year}{2019}).
\newblock


\bibitem[Thorne et~al\mbox{.}(2018)]%
        {thorne2018fever}
\bibfield{author}{\bibinfo{person}{James Thorne}, \bibinfo{person}{Andreas Vlachos}, \bibinfo{person}{Christos Christodoulopoulos}, {and} \bibinfo{person}{Arpit Mittal}.} \bibinfo{year}{2018}\natexlab{}.
\newblock \showarticletitle{FEVER: a large-scale dataset for fact extraction and VERification}.
\newblock \bibinfo{journal}{\emph{arXiv preprint arXiv:1803.05355}} (\bibinfo{year}{2018}).
\newblock


\bibitem[Touvron et~al\mbox{.}(2023)]%
        {touvron2023llama}
\bibfield{author}{\bibinfo{person}{Hugo Touvron}, \bibinfo{person}{Thibaut Lavril}, \bibinfo{person}{Gautier Izacard}, \bibinfo{person}{Xavier Martinet}, \bibinfo{person}{Marie-Anne Lachaux}, \bibinfo{person}{Timoth{\'e}e Lacroix}, \bibinfo{person}{Baptiste Rozi{\`e}re}, \bibinfo{person}{Naman Goyal}, \bibinfo{person}{Eric Hambro}, \bibinfo{person}{Faisal Azhar}, {et~al\mbox{.}}} \bibinfo{year}{2023}\natexlab{}.
\newblock \showarticletitle{Llama: Open and efficient foundation language models}.
\newblock \bibinfo{journal}{\emph{arXiv preprint arXiv:2302.13971}} (\bibinfo{year}{2023}).
\newblock


\bibitem[Wang et~al\mbox{.}(2023)]%
        {wang2023survey}
\bibfield{author}{\bibinfo{person}{Cunxiang Wang}, \bibinfo{person}{Xiaoze Liu}, \bibinfo{person}{Yuanhao Yue}, \bibinfo{person}{Xiangru Tang}, \bibinfo{person}{Tianhang Zhang}, \bibinfo{person}{Cheng Jiayang}, \bibinfo{person}{Yunzhi Yao}, \bibinfo{person}{Wenyang Gao}, \bibinfo{person}{Xuming Hu}, \bibinfo{person}{Zehan Qi}, {et~al\mbox{.}}} \bibinfo{year}{2023}\natexlab{}.
\newblock \showarticletitle{Survey on factuality in large language models: Knowledge, retrieval and domain-specificity}.
\newblock \bibinfo{journal}{\emph{arXiv preprint arXiv:2310.07521}} (\bibinfo{year}{2023}).
\newblock


\bibitem[Wei et~al\mbox{.}(2022)]%
        {wei2022chain}
\bibfield{author}{\bibinfo{person}{Jason Wei}, \bibinfo{person}{Xuezhi Wang}, \bibinfo{person}{Dale Schuurmans}, \bibinfo{person}{Maarten Bosma}, \bibinfo{person}{Fei Xia}, \bibinfo{person}{Ed Chi}, \bibinfo{person}{Quoc~V Le}, \bibinfo{person}{Denny Zhou}, {et~al\mbox{.}}} \bibinfo{year}{2022}\natexlab{}.
\newblock \showarticletitle{Chain-of-thought prompting elicits reasoning in large language models}.
\newblock \bibinfo{journal}{\emph{Advances in neural information processing systems}}  \bibinfo{volume}{35} (\bibinfo{year}{2022}), \bibinfo{pages}{24824--24837}.
\newblock


\bibitem[Xiao et~al\mbox{.}(2023)]%
        {xiao2023c}
\bibfield{author}{\bibinfo{person}{Shitao Xiao}, \bibinfo{person}{Zheng Liu}, \bibinfo{person}{Peitian Zhang}, {and} \bibinfo{person}{Niklas Muennighof}.} \bibinfo{year}{2023}\natexlab{}.
\newblock \showarticletitle{C-pack: Packaged resources to advance general chinese embedding}.
\newblock \bibinfo{journal}{\emph{arXiv preprint arXiv:2309.07597}} (\bibinfo{year}{2023}).
\newblock


\bibitem[Yan et~al\mbox{.}(2025)]%
        {yan2025unlocking}
\bibfield{author}{\bibinfo{person}{Bencheng Yan}, \bibinfo{person}{Shilei Liu}, \bibinfo{person}{Zhiyuan Zeng}, \bibinfo{person}{Zihao Wang}, \bibinfo{person}{Yizhen Zhang}, \bibinfo{person}{Yujin Yuan}, \bibinfo{person}{Langming Liu}, \bibinfo{person}{Jiaqi Liu}, \bibinfo{person}{Di Wang}, \bibinfo{person}{Wenbo Su}, {et~al\mbox{.}}} \bibinfo{year}{2025}\natexlab{}.
\newblock \showarticletitle{Unlocking Scaling Law in Industrial Recommendation Systems with a Three-step Paradigm based Large User Model}.
\newblock \bibinfo{journal}{\emph{arXiv preprint arXiv:2502.08309}} (\bibinfo{year}{2025}).
\newblock


\bibitem[Yang et~al\mbox{.}(2024)]%
        {yang2024qwen2}
\bibfield{author}{\bibinfo{person}{An Yang}, \bibinfo{person}{Baosong Yang}, \bibinfo{person}{Binyuan Hui}, \bibinfo{person}{Bo Zheng}, \bibinfo{person}{Bowen Yu}, \bibinfo{person}{Chang Zhou}, \bibinfo{person}{Chengpeng Li}, \bibinfo{person}{Chengyuan Li}, \bibinfo{person}{Dayiheng Liu}, \bibinfo{person}{Fei Huang}, {et~al\mbox{.}}} \bibinfo{year}{2024}\natexlab{}.
\newblock \showarticletitle{Qwen2 technical report}.
\newblock \bibinfo{journal}{\emph{arXiv preprint arXiv:2407.10671}} (\bibinfo{year}{2024}).
\newblock


\bibitem[Yasunaga et~al\mbox{.}(2022)]%
        {yasunaga2022deep}
\bibfield{author}{\bibinfo{person}{Michihiro Yasunaga}, \bibinfo{person}{Antoine Bosselut}, \bibinfo{person}{Hongyu Ren}, \bibinfo{person}{Xikun Zhang}, \bibinfo{person}{Christopher~D Manning}, \bibinfo{person}{Percy~S Liang}, {and} \bibinfo{person}{Jure Leskovec}.} \bibinfo{year}{2022}\natexlab{}.
\newblock \showarticletitle{Deep bidirectional language-knowledge graph pretraining}.
\newblock \bibinfo{journal}{\emph{Advances in Neural Information Processing Systems}}  \bibinfo{volume}{35} (\bibinfo{year}{2022}), \bibinfo{pages}{37309--37323}.
\newblock


\bibitem[Young et~al\mbox{.}(2024)]%
        {young2024yi}
\bibfield{author}{\bibinfo{person}{Alex Young}, \bibinfo{person}{Bei Chen}, \bibinfo{person}{Chao Li}, \bibinfo{person}{Chengen Huang}, \bibinfo{person}{Ge Zhang}, \bibinfo{person}{Guanwei Zhang}, \bibinfo{person}{Heng Li}, \bibinfo{person}{Jiangcheng Zhu}, \bibinfo{person}{Jianqun Chen}, \bibinfo{person}{Jing Chang}, {et~al\mbox{.}}} \bibinfo{year}{2024}\natexlab{}.
\newblock \showarticletitle{Yi: Open foundation models by 01. ai}.
\newblock \bibinfo{journal}{\emph{arXiv preprint arXiv:2403.04652}} (\bibinfo{year}{2024}).
\newblock


\bibitem[Yu et~al\mbox{.}(2022)]%
        {yu2022generate}
\bibfield{author}{\bibinfo{person}{Wenhao Yu}, \bibinfo{person}{Dan Iter}, \bibinfo{person}{Shuohang Wang}, \bibinfo{person}{Yichong Xu}, \bibinfo{person}{Mingxuan Ju}, \bibinfo{person}{Soumya Sanyal}, \bibinfo{person}{Chenguang Zhu}, \bibinfo{person}{Michael Zeng}, {and} \bibinfo{person}{Meng Jiang}.} \bibinfo{year}{2022}\natexlab{}.
\newblock \showarticletitle{Generate rather than retrieve: Large language models are strong context generators}.
\newblock \bibinfo{journal}{\emph{arXiv preprint arXiv:2209.10063}} (\bibinfo{year}{2022}).
\newblock


\bibitem[Zhang et~al\mbox{.}(2023)]%
        {zhang2023siren}
\bibfield{author}{\bibinfo{person}{Yue Zhang}, \bibinfo{person}{Yafu Li}, \bibinfo{person}{Leyang Cui}, \bibinfo{person}{Deng Cai}, \bibinfo{person}{Lemao Liu}, \bibinfo{person}{Tingchen Fu}, \bibinfo{person}{Xinting Huang}, \bibinfo{person}{Enbo Zhao}, \bibinfo{person}{Yu Zhang}, \bibinfo{person}{Yulong Chen}, {et~al\mbox{.}}} \bibinfo{year}{2023}\natexlab{}.
\newblock \showarticletitle{Siren's song in the AI ocean: a survey on hallucination in large language models}.
\newblock \bibinfo{journal}{\emph{arXiv preprint arXiv:2309.01219}} (\bibinfo{year}{2023}).
\newblock


\bibitem[Zhao et~al\mbox{.}(2023)]%
        {zhao2023felm}
\bibfield{author}{\bibinfo{person}{Yiran Zhao}, \bibinfo{person}{Jinghan Zhang}, \bibinfo{person}{I Chern}, \bibinfo{person}{Siyang Gao}, \bibinfo{person}{Pengfei Liu}, \bibinfo{person}{Junxian He}, {et~al\mbox{.}}} \bibinfo{year}{2023}\natexlab{}.
\newblock \showarticletitle{Felm: Benchmarking factuality evaluation of large language models}.
\newblock \bibinfo{journal}{\emph{Advances in Neural Information Processing Systems}}  \bibinfo{volume}{36} (\bibinfo{year}{2023}), \bibinfo{pages}{44502--44523}.
\newblock


\bibitem[Zhao et~al\mbox{.}(2024)]%
        {zhao2024felm}
\bibfield{author}{\bibinfo{person}{Yiran Zhao}, \bibinfo{person}{Jinghan Zhang}, \bibinfo{person}{I Chern}, \bibinfo{person}{Siyang Gao}, \bibinfo{person}{Pengfei Liu}, \bibinfo{person}{Junxian He}, {et~al\mbox{.}}} \bibinfo{year}{2024}\natexlab{}.
\newblock \showarticletitle{Felm: Benchmarking factuality evaluation of large language models}.
\newblock \bibinfo{journal}{\emph{Advances in Neural Information Processing Systems}}  \bibinfo{volume}{36} (\bibinfo{year}{2024}).
\newblock


\end{thebibliography}

\newpage
\appendix

\clearpage

\section{Negtive Sampling}
\label{negtive_sampling}
\begin{table}[h]
    \caption{An example of prompting LLMs to assist us with fine-grained filtering of negative candidates (Stage 3).}
    \label{tab:negtive_sampling}
    \centering
    \fontsize{9}{11}\selectfont
    \begin{tabular}{@{}p{8cm}@{}} 
        \toprule
        Given the *Question* and its *Answer*, please select three negative examples from the *Negative Candidates* according to the *Requirements* to create challenging multiple-choice distractors, following the *Output format*. Then provide reasoning. \cr
        
        \textbf{*Requirements*} \cr
        
        1. The negative examples should be inappropriate for completing the sentence in the *question* but possess effective confusion value that makes them hard to distinguish.
        
        2. The three negative examples should demonstrate diversity and cover multiple domains
        
        \textbf{*Output Format*} \cr
        Negative examples: Negative1, Negative2, Negative3 \cr
        
        \textbf{*Question*} \cr
        
        Please select the most suitable one from among the *Options* to fill in the blank space to complete the following *Sentence*, making it comply with the logic and knowledge in e-commerce and daily consumption. Just output the selected choice.
        The terms in brackets are the descriptions of the phrase. \cr
        Sentence: Versatile white (color) belongs to or is similar to \underline{\hspace{1cm}} (color) \cr
        
        \textbf{*Answer*} \cr
        Jasmine white \cr
        
        \textbf{*Negative Candidates*} \cr
        
        ['Pure white', 'Clean white', 'Cotton white', 'Cotton white', 'Classic white', 'Pure white', 'New white', 'Shell white', 'Cotton white', 'Vintage white', 'Translucent white', 'Greyish white', 'Versatile blue', 'Versatile grey', 'Versatile brown'] \cr
        \bottomrule
    \end{tabular}
\end{table}

\section{Verification}
\label{quality}
\begin{table}[h]
    \caption{An example of prompting LLMs to assist us in evaluating the question quality.}
    \label{tab:quality}
    \centering
    \fontsize{9}{11}\selectfont
    \begin{tabular}{@{}p{8cm}@{}} 
        \toprule
            Determine if the *Answer* satisfies the *Question*. First, answer yes or no, and then explain why. \cr
            \cr
            \textbf{*Question*:} Please select the most suitable one from among the *Options* to fill in the blank space to complete the following *Sentence*, making it comply with the logic and knowledge in e-commerce and daily consumption. Just output the selected choice. The terms in brackets are the descriptions of the phrase. \cr
            \textbf{*Sentence*}: Avène (brand) produces \underline{\hspace{1cm}} (name of category) \cr
            \textbf{*Options*}:  shower gel, lip balm, toner, serum \cr
            \textbf{*Answer*}: serum  \cr
        \bottomrule
    \end{tabular}
\end{table}

\section{Few-shot Prompt}
\label{fewshot_prompt}
\begin{table}[h]
    \caption{An example of few-shot question prompt.}
    \label{tab:fewshot_prompt}
    \centering
    \fontsize{9}{11}\selectfont
    \begin{tabular}{@{}p{8cm}@{}} 
        \toprule
        \textbf{*Question*:} Please select the most suitable one from among the *Options* to fill in the blank space to complete the following *Sentence*, making it comply with the logic and knowledge in e-commerce and daily consumption. Just output the selected choice. The terms in brackets are the descriptions of the phrase. Complete \#Example 3\# by imitating \#Example 1\# and \#Example 2\#. \cr
        \cr
        \textbf{\#Example 1\#} \cr
        *Sentence*: Rapid heating (function) is very important for \underline{\hspace{1cm}}  (name of category). \cr
        *Options*: Microwave oven, Flashlight, Electric fan, Speaker \cr
        *Answer*: Microwave oven \cr
        \cr
        \textbf{\#Example 2\#} \cr
        *Sentence*: Keeping warm (function) is important for \underline{\hspace{1cm}} (name of category) \cr
        *Options*: Short sleeve shirt, Down jacket, Slippers, Sun hat \cr
        *Answer*: Down jacket \cr
        \cr
        \textbf{\#Example 3\#} \cr
        *Sentence*: Avène (brand) produces \underline{\hspace{1cm}} (name of category) \cr
        *Options*:  shower gel, lip balm, toner, serum \cr
        *Answer*:   \cr
        \bottomrule
    \end{tabular}

\end{table}

\end{document}